\documentclass{article}

\usepackage{arxiv}

\usepackage{times}
\usepackage{latexsym} 
\usepackage{comment}
\usepackage{graphicx}
\usepackage{float}
\usepackage[inline]{enumitem}
\usepackage[T1]{fontenc}
\usepackage[utf8]{inputenc}
\usepackage{microtype}
\usepackage{natbib}
\usepackage{hyperref}       
\usepackage{url}            
\usepackage{booktabs}       
\usepackage{amsfonts}       
\usepackage{nicefrac}       

\newcount\Comments  
\usepackage{color}
\definecolor{darkgreen}{rgb}{0,0.5,0}
\definecolor{purple}{rgb}{1,0,1}
\newcommand{\kibitz}[2]{\ifnum\Comments=1\textcolor{#1}{#2}\fi}

\newif\ifworkingversion

\title{FinEAS: Financial Embedding Analysis of Sentiment}

\author{Asier Gutiérrez-Fandiño\\
  Barcelona Supercomputing Center / Barcelona\\
  LHF Labs\\
  \texttt{asier.gutierrez@bsc.es} \\
  \texttt{asier@lhf.ai}\\\And 
Miquel Noguer i Alonso \\
  Artificial Intelligence Finance Institute / New York City \\
  NYU Courant / New York City \\
  \texttt{miquel.noguer@aifinanceinstitute.com} \\\And
  Petter Kolm \\
  NYU Courant / New York City \\
  \texttt{petter.kolm@nyu.edu} \\\And
  Jordi Armengol-Estapé \\
  Barcelona Supercomputing Center / Barcelona \\
  LHF Labs\\
  \texttt{jordi.armengol@bsc.es} \\
  \texttt{jordi@lhf.ai}\\
  }

\begin{document}
\maketitle

\begin{abstract}



We introduce a new language representation model in finance called Financial Embedding Analysis of Sentiment (FinEAS).  
In financial markets, news and investor sentiment are significant drivers of security prices. Thus, leveraging the capabilities of modern NLP approaches for financial sentiment analysis is a crucial component in identifying patterns and trends that are useful for market participants and regulators.
In recent years, methods that use transfer learning from large Transformer-based language models like BERT, have achieved state-of-the-art results in text classification tasks, including sentiment analysis using labelled datasets. Researchers have quickly adopted these approaches to financial texts, but best practices in this domain are not well-established. 
In this work, we propose a new model for financial sentiment analysis based on supervised fine-tuned sentence embeddings from a standard BERT model. We demonstrate our approach achieves significant improvements in comparison to vanilla BERT, LSTM, and FinBERT, a financial domain specific BERT.

\end{abstract}

\section{Introduction}
\label{sec:intro}


Sentiment analysis is a technique where text is classified as conveying positive or negative meaning through the use of natural language processing (NLP). In the field of finance, news and investor sentiment are significant drivers of the individual security prices and the market as a whole. Natural Language Understanding (NLU) capabilities in the financial domain are needed for automating tasks and analysis.

Recently, practitioners of Natural Language Processing (NLP) who were involved in financial sentiment used a variety of cutting-edge machine learning algorithms, such as LSTMs. Researchers have quickly adopted modern NLP approaches to the financial domain, based on the latest successes in NLP, which leverage transfer learning from general Transformer-based language models. Nonetheless, there is a lack of knowledge about best practices for using Transformers in this domain. In this article, we propose an effective approach based on the use of Transformer language models that are explicitly developed for sentence-level analysis.

More specifically, we build upon the findings from Sentence-BERT \citep{reimers2019sentencebert}. In this work, the authors show that vanilla BERT does not provide strong ``out of the box'' sentence embeddings (unlike the token embeddings, which are either state-of-the-art or close to the state-of-the-art, depending on the task). Since financial sentiment is a sentence-level task, we base our approach on Sentence-BERT.

In Section \ref{sec:related_work}, we briefly review related work. Then, in Section \ref{sec:methods} we describe our proposed method. In Section \ref{sec:exp}, we present performance of our model and compare it to both common baselines and a domain-specific baseline. Finally, in Sections \ref{sec:discussion} and \ref{sec:conclusions} we summarize our results and conclusions. We make the code\footnote{\url{https://github.com/lhf-labs/finance-news-analysis-bert}} publicly available.

\section{Related Work}
\label{sec:related_work}

\paragraph{Background} Researchers and practitioners started using the lexicon-based bag-of-words, with a  predefined dictionary of positive and negative words. Here, one computes the sentiment score based on the number of matches of words in the text with each of the dictionaries. Indeed, this is exactly the approach taken by one of the first highly influential academic articles in the fields of text analysis in finance. In 2007, Paul Tetlock showed that the frequency of negative words in articles of the Wall Street Journal had predictive power of the future price moves of the Dow Jones Industrial Average Index and the daily volume traded on the New York Stock Exchange \citep{article}. Some work in the literature use a bag-of-words approach where the sentiment word lists are from the Loughran and McDonald financial dictionary (2009) \citep{LOUGHRAN:2011}. As a more advanced machine learning approach, researchers later introduced TF-IDF (Term Frequency–Inverse Document Frequency) for encoding text, and then trained supervised learning algorithms, such as SVMs (Support Vector Machines), with labelled datasets. An interesting application to Environmental, Social, and Governance (ESG)  and UN Sustainable Development Goals (SDG) investing can be found in \citet{Noguer2020}.

\paragraph{Current approaches} With the advent of deep learning and its application in NLP, researchers began applying Recurrent Neural Networks (RNNs) and Convolutional Neural Networks (CNNs) for text classification. The current state-of-the-art in text classification typically involves a purely attentional architecture, the Transformer architecture \citep{DBLP:journals/corr/VaswaniSPUJGKP17}, especially by fine-tuning pre-trained models. Specifically for financial sentiment, we highlight \citet{Araci2019FinBERTFS}, in which the authors fine-tune a pre-trained Transformer in the Financial Phrasebank dataset \citep{Malo2014GoodDO}.


\paragraph{BERT} Bidirectional Encoder Representations from Transformers (BERT) \citep{DBLP:journals/corr/abs-1810-04805} is a Transformer encoder pre-trained on large textual corpora without supervision. The attention mechanism of the Transformer allows obtaining \textit{contextual} word embeddings, i.e. word embeddings taking into account the rest of the word embeddings in the sentence. For pre-training without supervision, BERT uses two surrogate tasks:
\begin{itemize}
    \item Masked Language Modeling (MLM): 15\% of the tokens are randomly \textit{masked} (i.e., replaced with the special token \texttt{<MASK>}, and the model is asked to predicted them. To do so, the model must learn useful representations from the context. In this way, it learns to produce token-level embeddings.
    \item Next Sentence Prediction (NSP): Each training instance consists of a sentence pair. Half of the time, the second sentence is a random sentence; in the other 50\% of occurrences, the second sentence is the actual sentence that appears next to the original sentence. The model must predict, from the embeddings of the special token \texttt{<CLS>} (class), whether the second sentence is the next one or not. In this fashion, it learns to produce sentence-level embeddings.
\end{itemize}

\paragraph{Domain-specific models} Several studies in different domains have demonstrated that domain-specific BERT models can outperform the generic BERT. Perhaps most well-known are BioBERT \citep{DBLP:journals/corr/abs-1901-08746} and SciBERT \citep{Beltagy2019SciBERT} in the biomedical/scientific domain. In the case of the financial domain, starting from the original English BERT, FinBERT \citep{Araci2019FinBERTFS} was fine-tuned on the Financial Phrasebank \citep{Malo2014GoodDO} and FiQA Task 1 sentiment scoring dataset,\footnote{\url{https://sites.google.com/view/fiqa}} thereby achieving state-of-the-art results. 

\paragraph{Sentence-BERT} In \citet{reimers2019sentencebert}, authors noted that the sentence embeddings obtained from vanilla BERT (the ones pre-trained with the NSP task) lack in quality. In fact, considerably simpler baselines are competitive with BERT in this regard (e.g., averaging word embeddings). \citet{DBLP:journals/corr/abs-1907-11692} emphasized that the NSP task was not as useful as thought, and authors suggested removing it from the BERT pre-training scheme. Consequently, \citet{reimers2019sentencebert} propose the Sentence-BERT model. Starting from a pre-trained BERT checkpoint, they fine-tune it with supervision with a Siamese BERT network (meaning that they encode pairs of sentences with the same encoder), and predict the sentence entailment from the two sentence embeddings (Natural Language Inference (NLI) task). This approach results in more meaningful sentence representations.

It is now well-known that pre-trained Transformers achieve state-of-the-art performance in NLP tasks \citep{Araci2019FinBERTFS}. In this article, unlike \citet{Araci2019FinBERTFS}, rather than starting from vanilla BERT, which is state-of-the-art for token-level embeddings but not for sentence-level tasks, we base our work on a model that has been fine-tuned for producing high-quality sentence embeddings. We believe this is a more sensible approach in the case of financial sentiment analysis. In contrast to  \citet{Araci2019FinBERTFS}, we model financial sentiment as a continuous variable (from -1 to 1), instead of using discrete values.

\section{Methods}
\label{sec:methods}

We start from two observations:
\begin{enumerate}
    \item The financial domain is considerably similar in lexicon and structure to the general domain.
    \item Financial sentiment analysis is a sentence/document-level task.
\end{enumerate}

\paragraph{Domain} Based on the first observation, we conjecture that a domain-specific BERT model, even if presumably optimal for this task, might not be worth the effort in terms of the compute time and large amounts of training data needed. Instead, we suggest using a general-domain model as the NLP backbone.

\paragraph{Sentence-level} Regarding the second observation, while financial sentiment does require high-quality sentence embeddings (not token-level embeddings), we note that vanilla BERT does not provide strong sentence embeddings.

\paragraph{Approach} We propose a new model that starts from supervised fine-tuned sentence embeddings from a standard BERT model. Specifically, we feed the sentences to the Sentence-BERT model, and then we try both using it as a feature extractor and perform full-model fine-tuning. The output sentence embedding, with a dimension of 768, is fed to a linear layer attached to a \texttt{tanh} activation function (since the task is a regression between -1 and 1). We refer to the new model as Financial Embedding Analysis of Sentiment (FinEAS).

\section{Experiments and Results}
\label{sec:exp}

\paragraph{Data} To train and test our models we use a large-scale financial analysis news dataset \texttt{US\_news} from RavenPack.\footnote{\url{https://www.ravenpack.com}} RavenPack Analytics delivers sentiment scores and event-based data that are likely to have an impact on security prices and financial markets worldwide. The service includes analytics for more than 300,000 entities in over 130 countries and covers over 98\% of the investable global market.
Apart from its scale and scope, comprising an extensive period, it also has the advantage of being easily filtered and sampled via the tools offered by RavenPack. In our analysis, we use three samples of this dataset:
\begin{enumerate}
    \item Six months of data leading up to February 11, 2021, resulting in 2,279,823 instances;
    \item One year of data leading up to February 11, 2021, resulting in 4,847,629 instances; and 
    \item Two years of data leading up to February 11, 2021, resulting in 12,358,024 instances.
\end{enumerate}
In addition, we use the following two weeks (February 12, 2021 through February 26, 2021) as out-of-sample data for testing whether models exhibits predictive power in out-of-distribution settings (time shift). This sample consists of 274,190 instances. Figure \ref{fig:company_test} shows that, even if relatively similar, the company distribution in the 2-year and additional test sets are, indeed different.

For all samples, we apply the following filters:
\begin{enumerate}
    \item We filter by companies (column \texttt{COMP}), retaining the top fifty companies in the US.\footnote{As per RavenPack criteria. The companies are all publicly listed.}
    \item We remove any duplicate entries.
\end{enumerate}
For the model, we use the free text from \texttt{EVENT\_TEXT} (consisting of the headline) as input and \texttt{EVENT\_SENTIMENT\_SCORE} (-1 to +1) as the target. We randomly split the samples into train, validation, and test sets, with a proportion of 99.5-0.25-0.25.

\begin{figure*}[!htb]
\centering
    \includegraphics[scale=0.25]{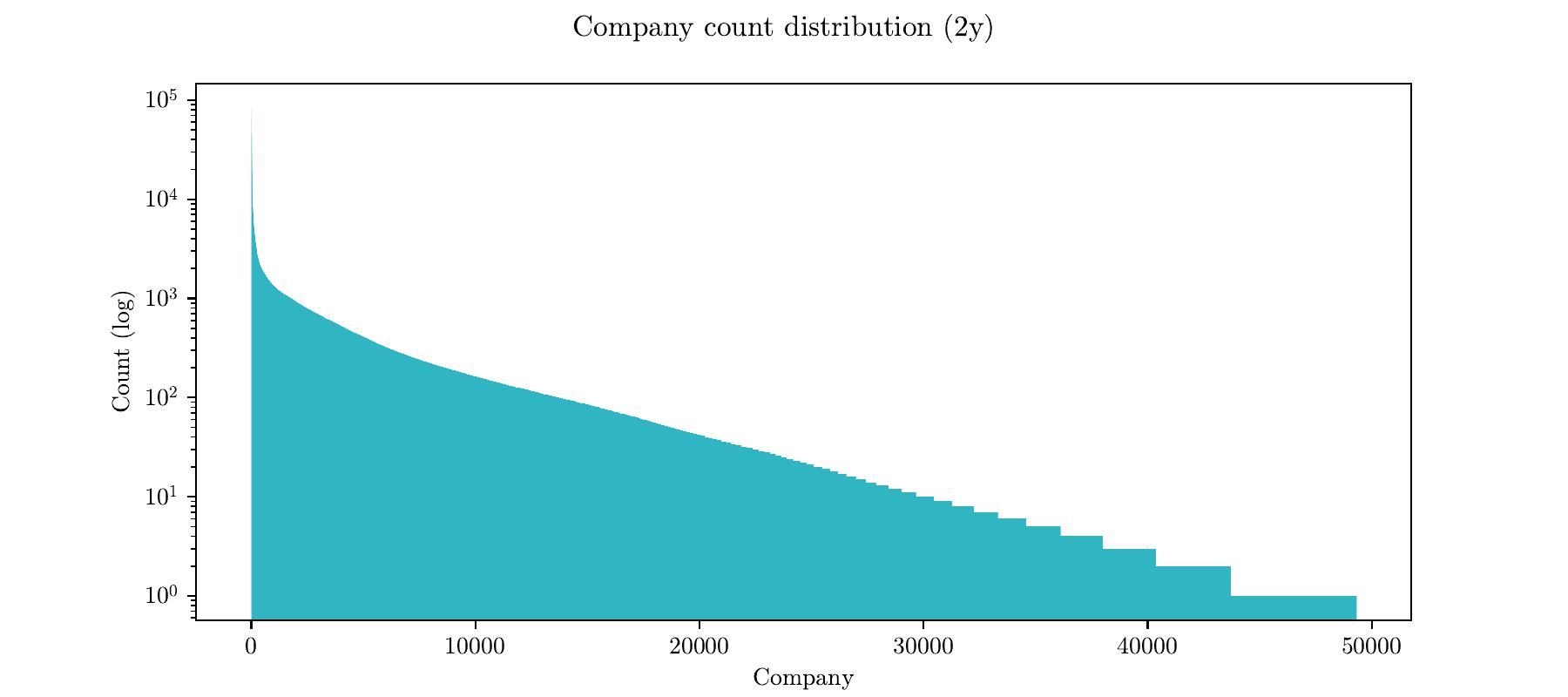}
    
    \includegraphics[scale=0.25]{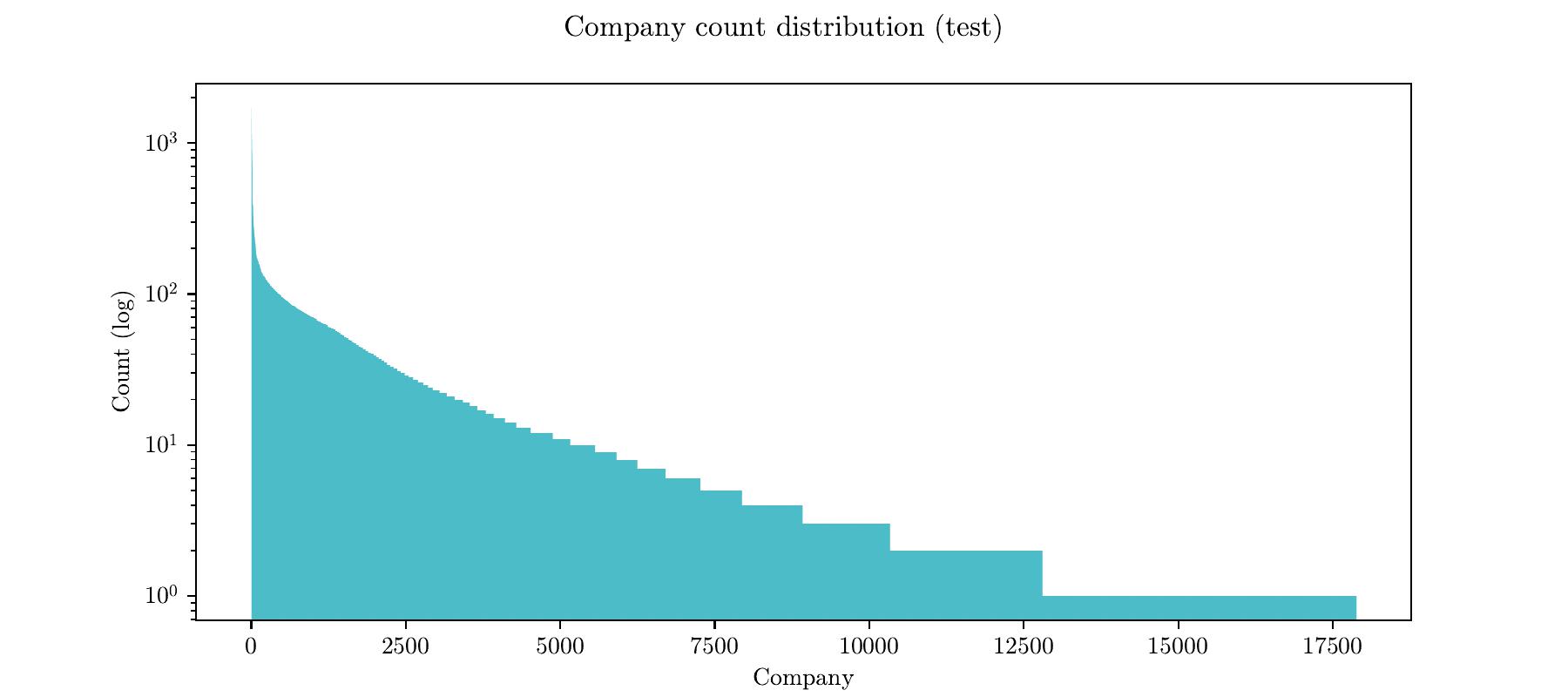}

    \caption{Company distribution for the 2-year and additional test sets.}
    \label{fig:company_test}
\end{figure*}

\paragraph{Implementation} We implement the model in PyTorch \cite
{NEURIPS2019_9015} and use the official Sentence-BERT implementation. 
 
\paragraph{Experimental framework} For evaluating our approach, we compare the following models: 
\begin{enumerate}
    \item FinEAS: A Sentence-BERT (base) with an additional linear layer for the regression.
    \item FinBERT \citep{Araci2019FinBERTFS}.
    \item A BERT (base) with an additional linear layer for the regression.
    \item A Bi-directional Long-Short Term Memory (LSTM) \citep{10.1162/neco.1997.9.8.1735} network with an additional linear layer for the regression. For the LSTM, we use two layers, with a hidden size of 256, and a dropout rate of 0.2.  
\end{enumerate}
First, we compare our approach with BERT and an LSTM. For this initial experiment, for both BERT and FinEAS, we freeze the weights of the model, and use the models as feature extractors. Then, we compare FinEAS with full-model fine-tuning with FinBERT (also with full-model fine-tuning).

For all models, we employ the same amount of maximum number of epochs, and early stopping based on the monitoring of the validation loss with a patience of 5. Also, we choose a batch size of 32 sentences and use the Adam \citep{kingma2017adam} optimizer with a learning rate of 0.001. We train all models on an NVIDIA GPU. Regarding the tokenization, for BERT, FinBERT and FinEAS we use their subword-based, pre-trained Wordpiece \citep{DBLP:journals/corr/WuSCLNMKCGMKSJL16} tokenizers. For the LSTM, we use the word-based tokenizer from Spacy \citep{spacy}. All models are evaluated on the same evaluation splits using Mean Squared Error (MSE) loss.

\paragraph{Results} Table \ref{tbl:results1} shows the results for the initial comparison, that is, BERT and FinEAS with the backbone frozen vs.~a fully trained LSTM. FinEAS achieves large improvements in MSE with a value of 0.0556 compared to a BiLSTM (0.2108) and a BERT baseline (0.2124) for 6 months. For the other time frames, we observe a similar relative and absolute performance of FinEAS to the other approaches. This provides support for that our results are not  artifacts of a specific sample or model overfitting. In particular, we emphasize that in each two week test set, FinEAS shows remarkable gains with respect to the baselines. 

\begin{table}
    \centering
    \begin{tabular}{|l|c|c|c|}
    \hline
         & FinEAS & BERT & BiLSTM \\ \hline
        6 months & \textbf{0.0556} & 0.2124 & 0.2108 \\ \hline
        $\hookrightarrow$ next 2w & \textbf{0.1061} & 0.2190 & 0.2194 \\ \hline
        12 months & \textbf{0.0654} & 0.2137 & 0.2140 \\ \hline
        $\hookrightarrow$ next 2w & \textbf{0.1058} & 0.2191 & 0.2194 \\ \hline
        24 months & \textbf{0.0671} & 0.2087 & 0.2086 \\ \hline
        $\hookrightarrow$ next 2w & \textbf{0.1065} & 0.2188 & 0.2185 \\ \hline
    \end{tabular}
    \caption{Initial experiments: MSE for the FinEAS, BERT and BiLSTM models for different subsets of the RavenPack dataset. Here, we kept the backbone models for FinEAS and BERT frozen during training.}
    \label{tbl:results1}
\end{table}

Table \ref{tbl:results2} shows the final comparison, once the first one has shown that FinEAS outperforms the basic baselines. In this one, FinEAS is compared to FinBERT, a state-of-the-art model for financial sentiment analysis. Both models are fully trained with no models frozen. While FinBERT has been trained on financial sentiment analysis data previously to our fine-tuning, FinEAS starts from scratch in the sense that it has never been specifically fine-tuned to the financial domain. Nonetheless, FinEAS outperforms FinBERT in the three temporal scenarios.

\begin{table}
    \centering
    \begin{tabular}{|l|c|c|}
    \hline
         & FinEAS & FinBERT  \\ \hline
        6 months & \textbf{0.0044} & 0.0050  \\ \hline
        12 months & \textbf{0.0036} & 0.0034  \\ \hline
        24 months & \textbf{0.0033} & 0.0040  \\ \hline
    \end{tabular}
    \caption{MSE for the FinEAS and FinBERT models for different subsets of the RavenPack dataset. None of the models are frozen during training.}
    \label{tbl:results2}
\end{table}

\section{Discussion}
\label{sec:discussion}

\paragraph{Results} FinEAS, our proposed approach, clearly outperforms two common baselines, the vanilla BERT and a bidirectional LSTM, and also obtains better results than FinBERT, a financial domain specific BERT. Perhaps somewhat surprisingly, the vanilla BERT does not outperform the bidirectional LSTM, suggesting its sentence embeddings may not be good enough for this kind of text. The results are consistent across time frames, and our model shows robust out-of-sample performance on a two-week holdout set. 

\paragraph{Sentence embeddings} The poor results of the vanilla BERT supports the hypothesis that the sentence embeddings learned with the NSP task are not suitable for the financial sentiment task (at least not ``out of the box''), which requires a high degree of sentence-level understanding. Supervised Sentence-BERT fine-tunes the sentence embeddings of BERT in a general-domain dataset. This is coherent with the intuition that for financial sentiment the most important aspect is the quality of the sentence-level embeddings, not the specific structure or vocabulary in the financial domain.



\paragraph{Limitations of our study} We use a single dataset in our study. As it is a commercial dataset, it is not accessible to the general public. However, the dataset is large, diverse and of high quality. 


\section{Conclusion and Future Work}
\label{sec:conclusions}

We have demonstrated that FinEAS, a model based on BERT pre-trained on the general domain but fine-tuned for sentence-level tasks, is a sensible approach for financial sentiment classification. In conclusion, our model is simple to implement and outperforms several common baselines, including vanilla BERT and task-specific approaches. We make our code and model weights publicly available. In future work, we think it will be interesting to further explore Transformers in the financial domain, with an emphasis on models fine-tuned for sentence and/or document-level tasks.


\clearpage
\bibliography{anthology,custom}
\bibliographystyle{apalike} 



\clearpage
\section*{Appendix}

\begin{figure}[H]
\centering
    \includegraphics[scale=0.2]{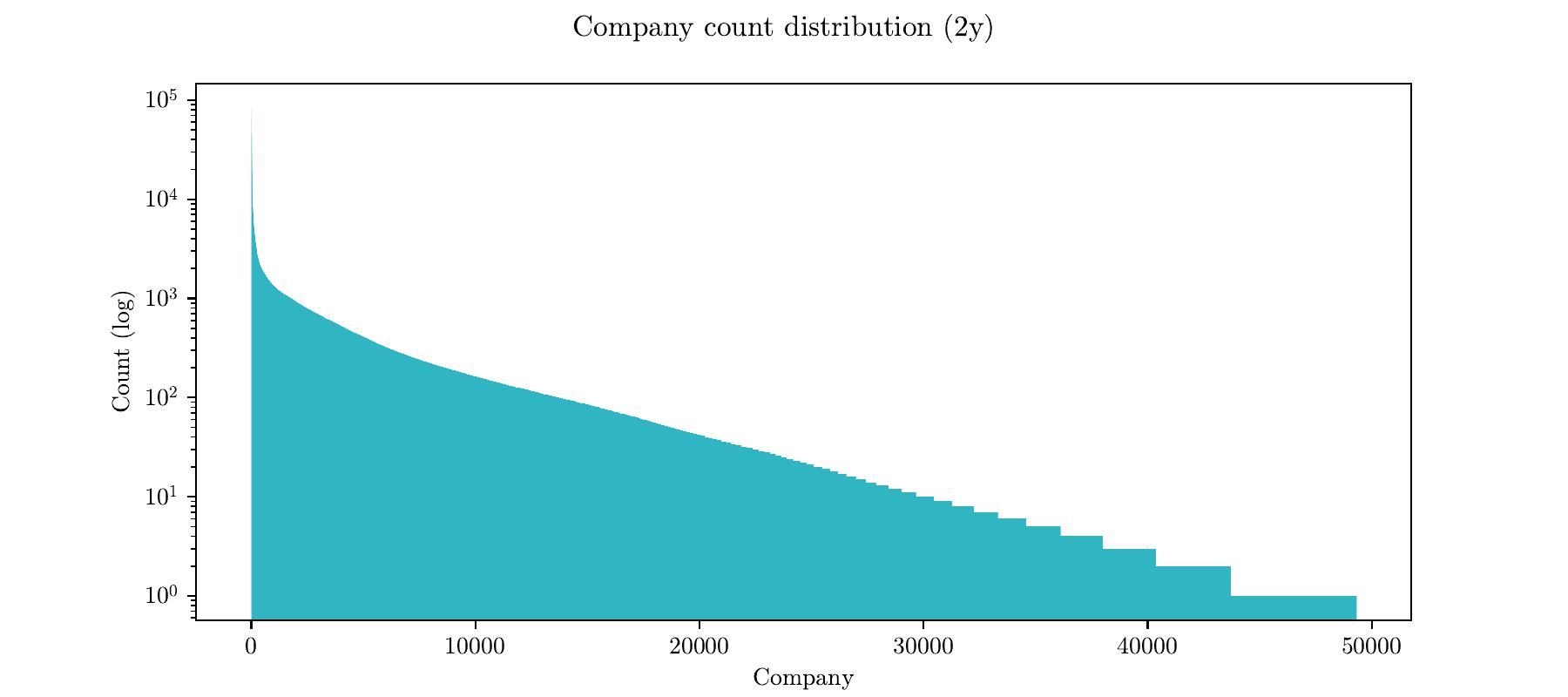}

    \caption{Company distribution for the 2-years dataset.}
    \label{fig:company_2y}
\end{figure}
\begin{figure}[H]
\centering
    \includegraphics[scale=0.2]{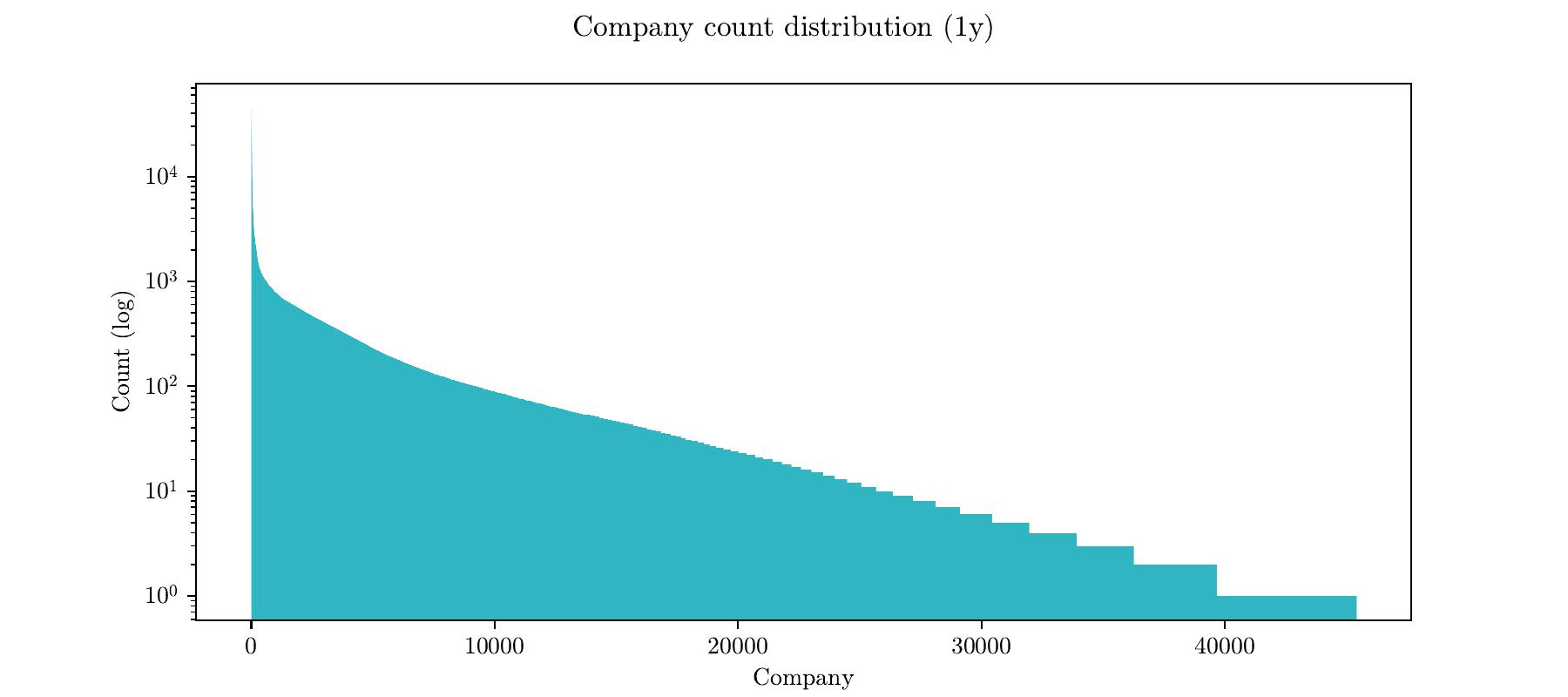}

    \caption{Company distribution for the 1-year dataset.}
    \label{fig:company_1y}
\end{figure}
\begin{figure}[H]
\centering
    \includegraphics[scale=0.2]{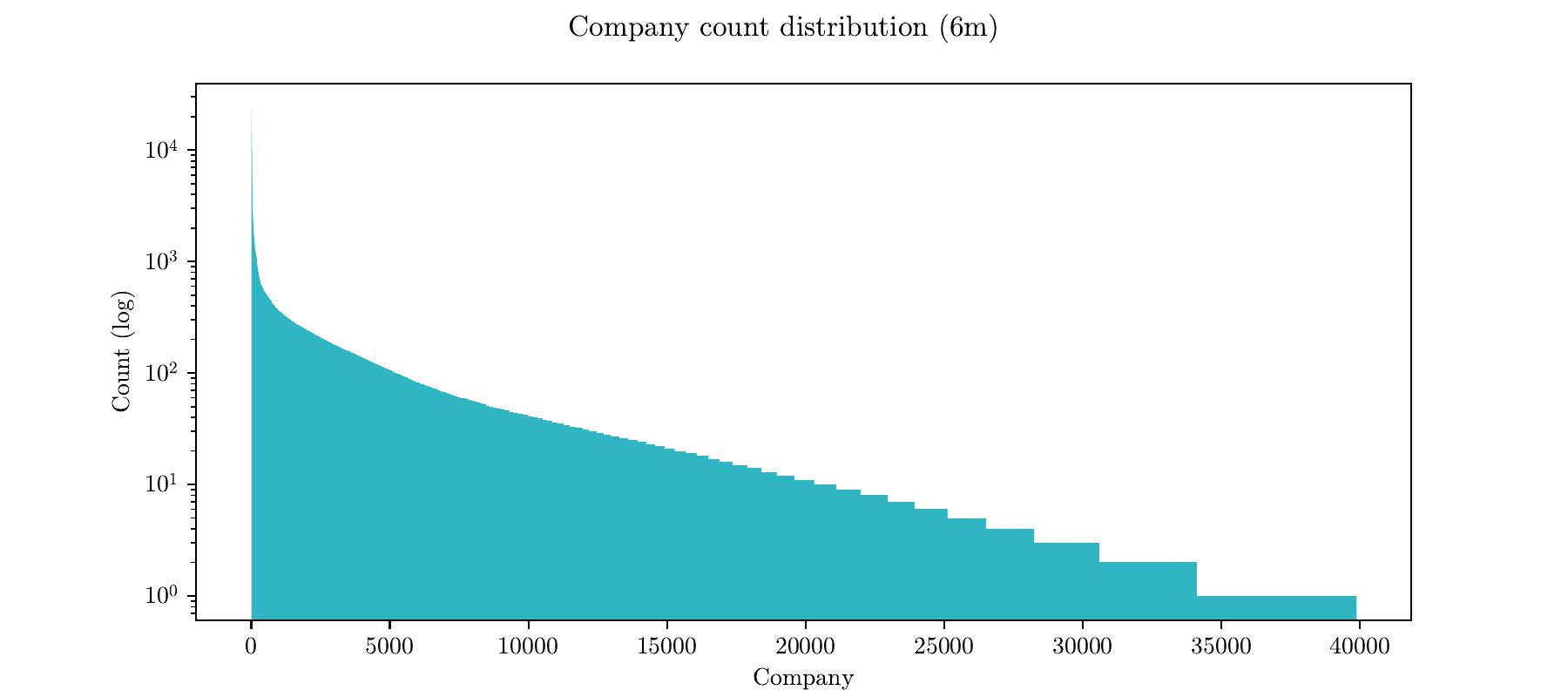}

    \caption{Company distribution for the 6-months dataset.}
    \label{fig:company_6m}
\end{figure}

\begin{figure}[H]
\centering
    \includegraphics[scale=0.2]{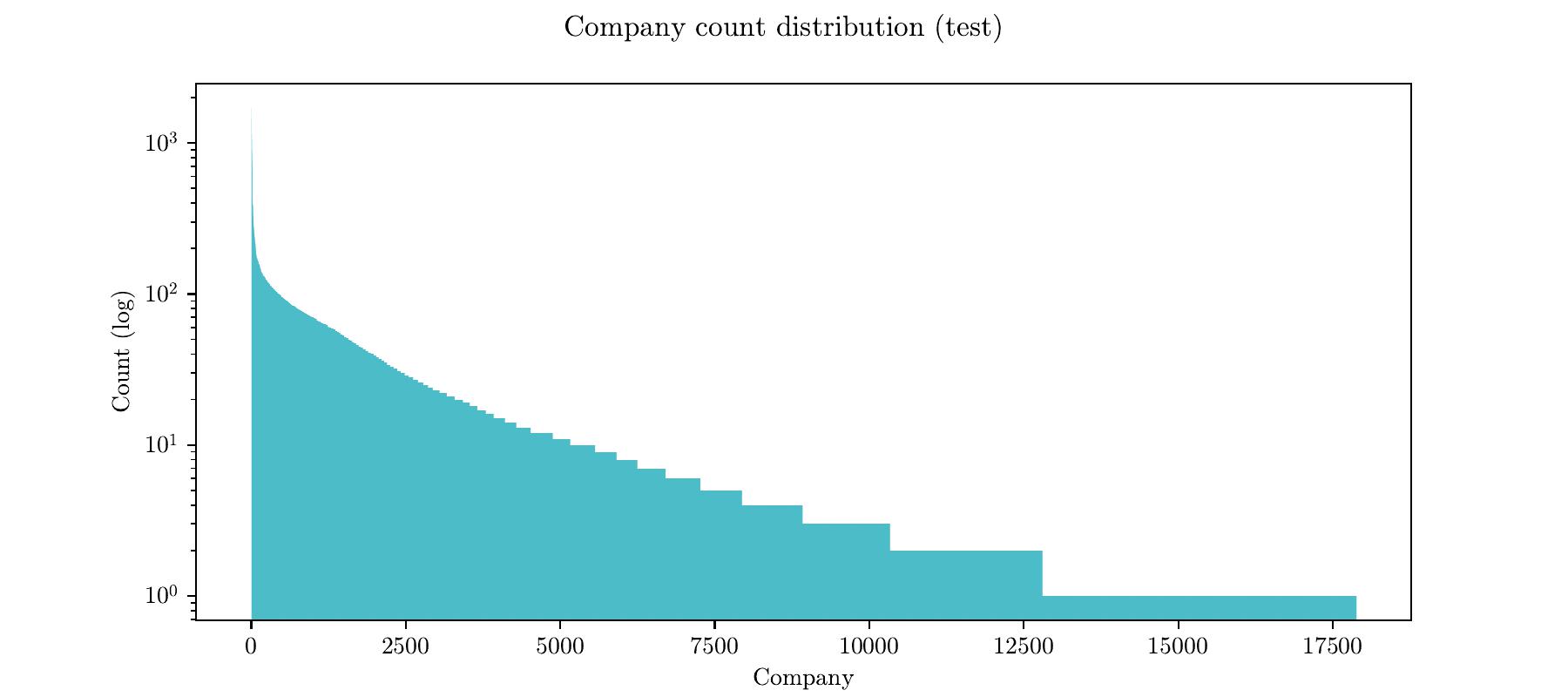}

    \caption{Company distribution for the test set.}
    \label{fig:company_test}
\end{figure}

\begin{figure}[H]
\centering
    \includegraphics[scale=0.2]{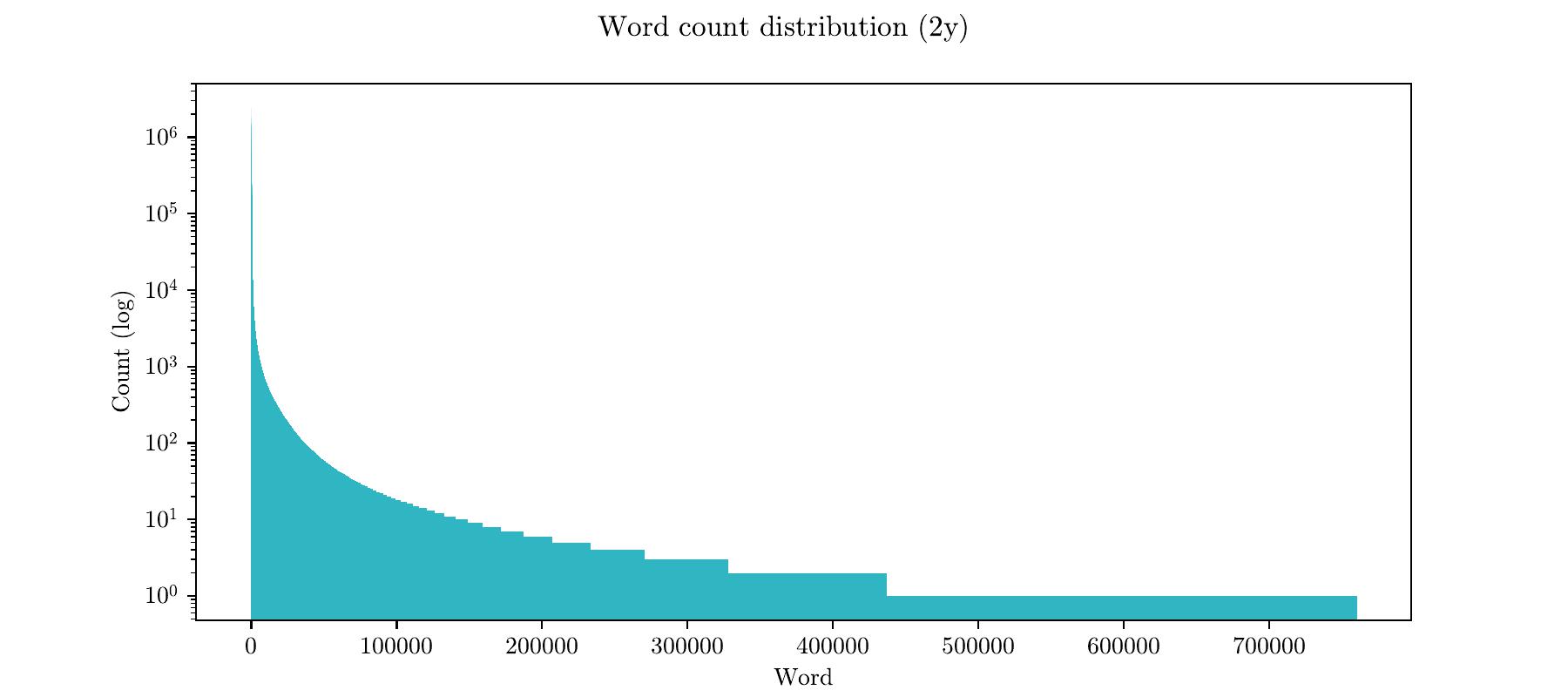}

    \caption{Word count distribution for the 2-years dataset.}
    \label{fig:word_2y}
\end{figure}
\begin{figure}[H]
\centering
    \includegraphics[scale=0.2]{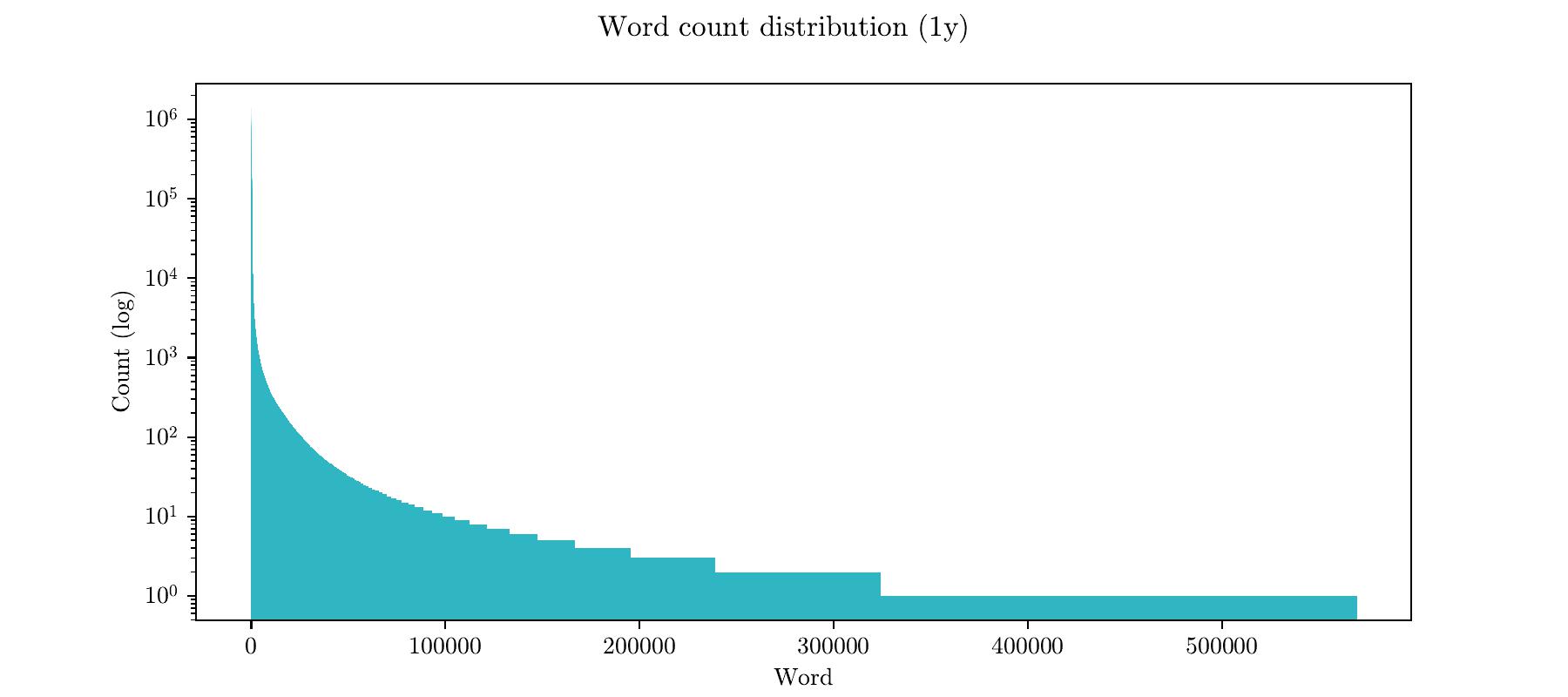}

    \caption{Word count distribution for the 1-year dataset.}
    \label{fig:word_1y}
\end{figure}
\begin{figure}[H]
\centering
    \includegraphics[scale=0.2]{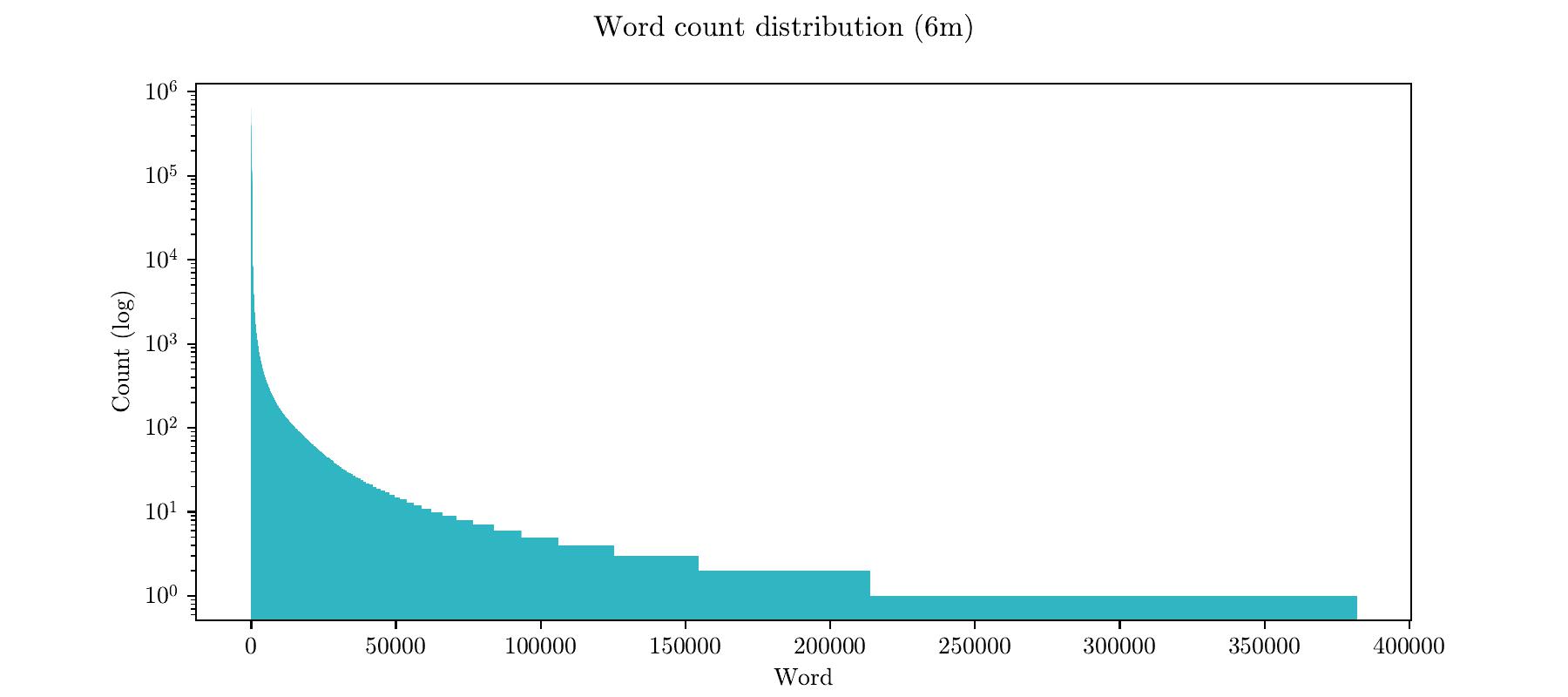}

    \caption{Word count distribution for the 6-months dataset.}
    \label{fig:word_6m}
\end{figure}

\begin{figure}[H]
\centering
    \includegraphics[scale=0.2]{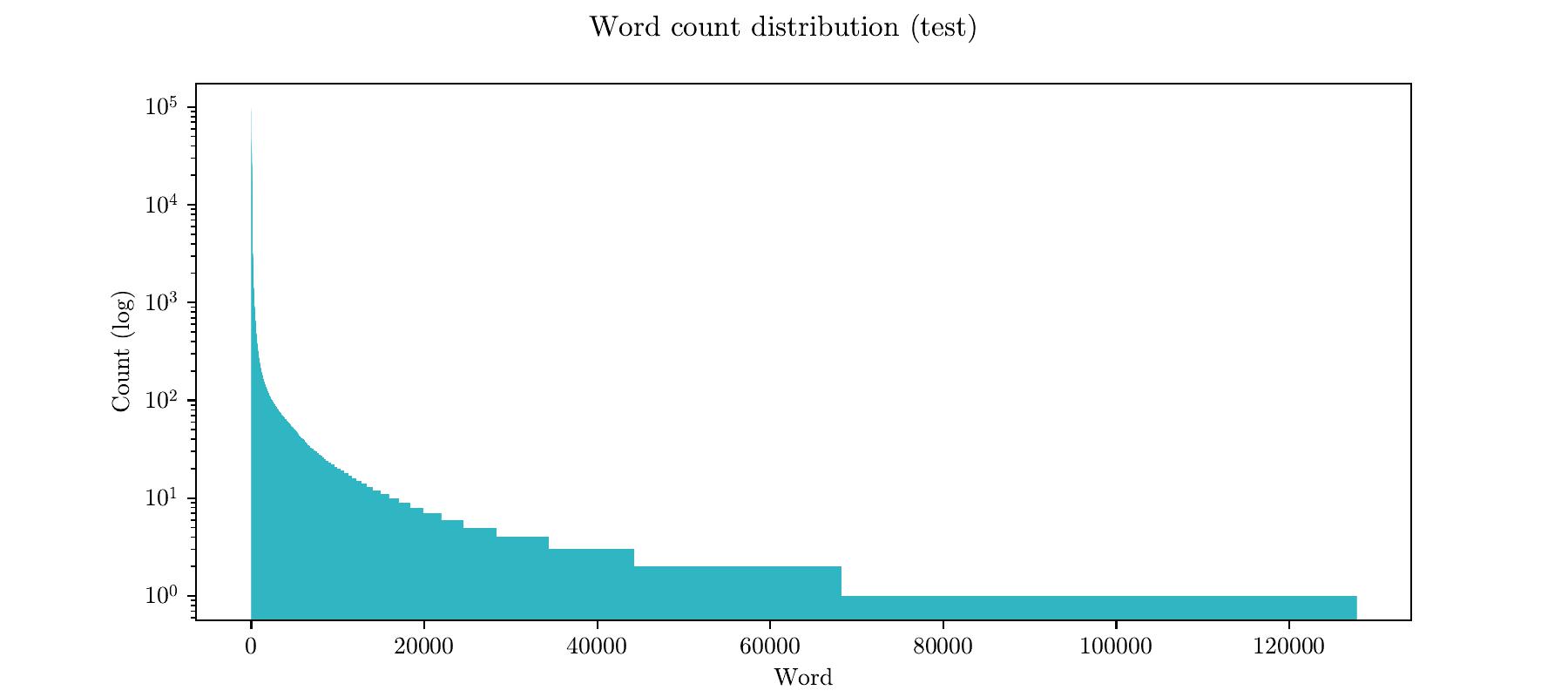}

    \caption{Word count distribution for the test set.}
    \label{fig:word_test}
\end{figure}

\begin{figure}[H]
\centering
    \includegraphics[scale=0.2]{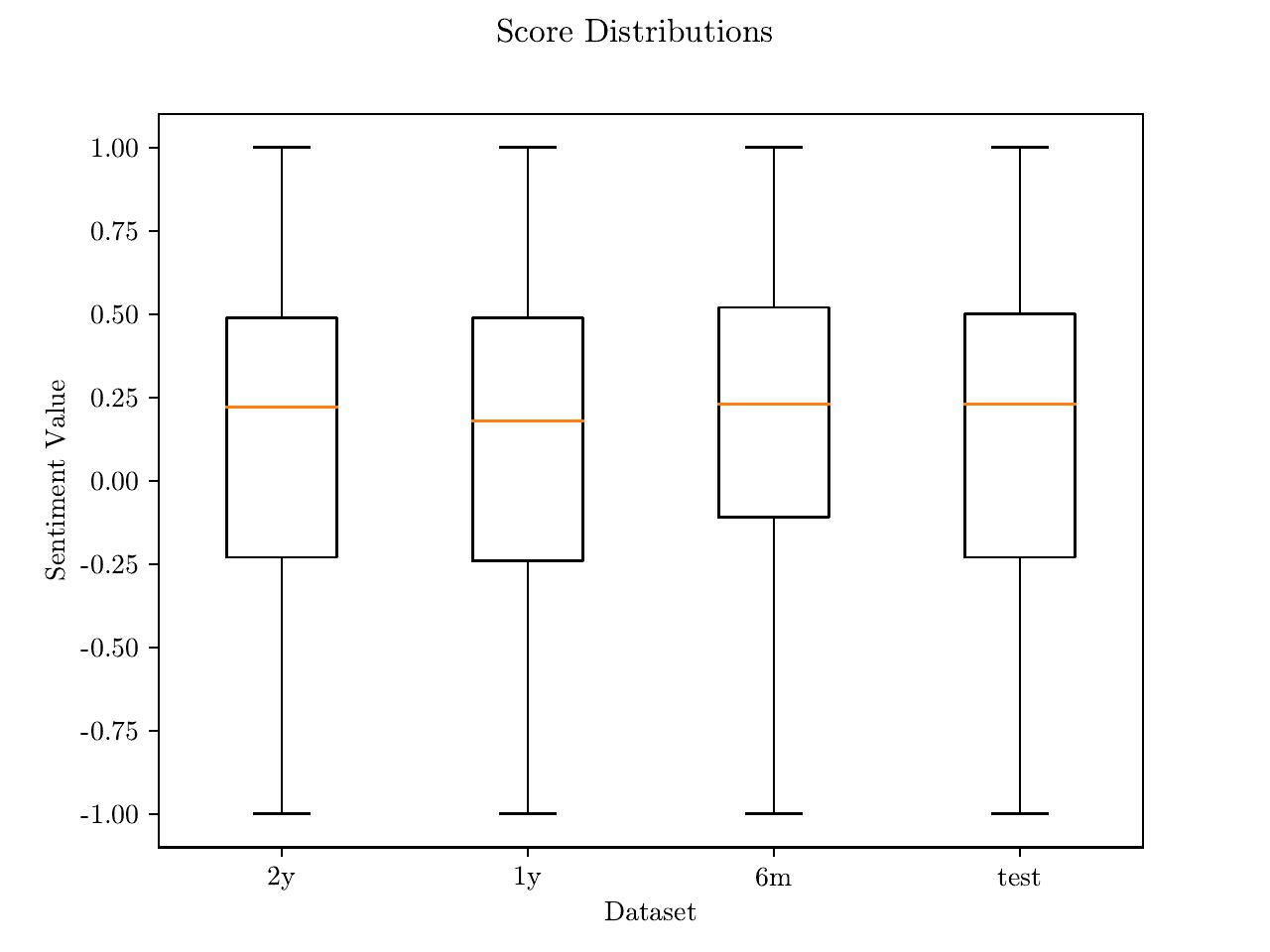}

    \caption{Sentiment scores distribution for the different data splits.}
    \label{fig:scores}
\end{figure}

\end{document}